\documentclass[runningheads]{llncs}
\usepackage{graphicx}
\usepackage{amsmath}
\usepackage{color,xcolor}
\usepackage{amssymb}
\usepackage{stfloats}

\usepackage{booktabs}       
\usepackage{siunitx}        
\usepackage{xcolor}          
\usepackage{multirow}
\usepackage{graphicx}

\usepackage{hyperref}
\hypersetup{hypertex=true,
            colorlinks=true,
            linkcolor=blue,
            anchorcolor=blue,
            citecolor=blue}

\begin{document}
\title{STAR-Pose: Efficient Low-Resolution Video Human Pose Estimation via Spatial-Temporal Adaptive Super-Resolution}
\titlerunning{Efficient Low-Resolution Video Human Pose Estimation}
%
\author{Yucheng Jin\inst{1} \and
Jinyan Chen\inst{2} \and
Ziyue He\inst{3} \and
Baojun Han\inst{4} \and
Furan An\inst{5}
}
\authorrunning{Jin et al.}
%
\institute{
Tianjin University \\
\email{chenjinyan@tju.edu.cn} 
}%
\maketitle              
\begin{abstract}
Human pose estimation in low-resolution videos presents a fundamental challenge in computer vision. Conventional methods either assume high-quality inputs or employ computationally expensive cascaded processing, which limits their deployment in resource-constrained environments. We propose STAR-Pose, a spatial-temporal adaptive super-resolution framework specifically designed for video-based human pose estimation. Our method features a novel spatial-temporal Transformer with LeakyReLU-modified linear attention, which efficiently captures long-range temporal dependencies. Moreover, it is complemented by an adaptive fusion module that integrates parallel CNN branch for local texture enhancement. We also design a pose-aware compound loss to achieve task-oriented super-resolution. This loss guides the network to reconstruct structural features that are most beneficial for keypoint localization, rather than optimizing purely for visual quality. Extensive experiments on several mainstream video HPE datasets demonstrate that STAR-Pose outperforms existing approaches. It achieves up to 5.2\% mAP improvement under extremely low-resolution (64×48) conditions while delivering 2.8× to 4.4× faster inference than cascaded approaches.
\end{abstract}
\section{Introduction}
Human Pose Estimation (HPE) is a fundamental area in computer vision with applications in human-computer interaction, action recognition, and intelligent surveillance. In practice, input videos are frequently constrained by environmental factors and camera limitations, resulting in human instances appearing as only a few pixels. This severely degrades the performance of existing HPE methods. Mainstream Heatmap-based approaches\cite{HRNet} map continuous coordinates to discrete heatmap representations, where low-resolution input amplifies inevitable quantization errors, causing a sharp decline in joint localization accuracy.

Current solutions focus on optimizing the heatmap generation process\cite{TinyPeoplePose,SRHeatMap} or post-processing\cite{UDP}. Howerver, these approaches are primarily designed for single-frame image and ignore the abundant temporal information present in video sequences, yielding limited improvements under low-resolution conditions.

Video Super-Resolution (V-SR) leverages complementary information across frames to recover missing details. However, cascading generic V-SR methods\cite{SR19,SR21} with HPE introduces two issues: (1) These methods focus on pixel-level optimization which may produce artifacts or excessive smoothing that degrades human body representaion; (2) The two-stage pipeline introduces significant computational overhead. These limitations motivate the need for an efficient, task-specific video super-resolution approach dedicated to HPE.

To address this, we propose STAR-Pose, which innovatively integrates task-aware video super-resolution with HPE. We employ LeakyReLU-modified linear attention\cite{LinearTransformer1}, reducing the computational complexity from $\mathcal{O}(N^2)$ to $\mathcal{O}(N)$ for efficient long-range temporal modeling. Parallel CNN branches enhance feature extraction capability via multi-scale processing. A pose-guided compound loss directs super-resolution processing toward reconstructing structural information beneficial for joint localization. Our contributions are:

\begin{itemize}
    \item We introduce \textbf{STAR-Pose}, an efficient spatial-temporal architecture combining linear attention-based Transformer with parallel CNN branch, achieving $\mathcal{O}(N)$ complexity for temporal modeling and local detail extraction.
\end{itemize}
\begin{itemize}
    \item We employ a task-driven optimization strategy by incorporating pose supervision into SR training, focusing the reconstruction on human structure-relevant features to improve joint prediction accuracy.
\end{itemize}
\begin{itemize}
    \item A comprehensive evaluation across benchmark datasets under different low-resolution scenarios, demonstrating superior performance with lower computational overhead compared to SOTA HPE methods and SR+HPE cascades.
\end{itemize}
\section{Related Work}

\subsection{Video Super-Resolution}
Video super-resolution exploits temporal information to recover high-resolution details from low-resolution sequences. CNN methods\cite{SR2} learned direct mappings from low to high resolution, were later extended to video through inter-frame relationships modeling via recursive structures and deformable convolutions\cite{SR12,SR17}. GAN-based approaches\cite{SR6_2} employ adversarial learning for realistic texture generation. Recent Transformer methods\cite{SR8,SR13,SR14}excel at capturing temporal correlations through long-range dependency modeling, achieving superior detail restoration but incurring quadratic computational complexity in long sequence data such as video\cite{SR16,SR21}.

Task-driven SR methods\cite{TaskSR2} incorporate downstream supervision supervisory signals into the training process, enabling the reconstructed images to better serve specific applications. For human-related tasks, some works\cite{HumanSR3} utilize human parsing maps and texture priors to enhance details in key regions. Despite these advances, HPE-supporting SR research remains limited to static images, underutilizing temporal information in videos.

\subsection{Human Pose Estimation in Low-Resolution}
Human pose estimation localizes human keypoints from image/video inputs.  The main challenges of low-resolution scenarios include: (1) recognizing small-scale human instances with blurred details; (2) amplified quantization errors caused by heatmap representations at low resolution. Existing solutions mainly encompass optimizing the generation and decoding process of heatmaps. \cite{HRNet} maintaining high-resolution feature representations; \cite{SRHeatMap} treats heatmap generation as a super-resolution task; \cite{TinyPeoplePose} refines heatmaps via Gaussian distributions. Post-processing approaches like \cite{UDP} employ unbiased data transformation and decoding strategies. Recently, some regression methods\cite{DSTA} have been utilized to avoid quantization error issues.

These approaches mitigate feature-level quantization errors, but useful features extracted from inherently blurred low-resolution images inputs remain limited. Some strategies integrate super-resolution into HPE: \cite{SRPose3} adds SR modules at image and feature levels; \cite{SRPose4} jointly trains SR, HPE and action recognition; \cite{SRPose2} incorporates pose estimation loss into SR training. While effective, these methods target single-frame images, ignoring temporal continuity in video. Although some video HPE methods\cite{DCPose,DSTA} capture motion cues, they assume high-quality input, limiting their performance under low-resolution conditions.

we propose a end-to-end framework specifically designed for low-resolution video HPE, utilizing improved linear attention for efficient spatial-temporal modeling across multiple frames and pose-guided SR focused on human structure reconstruction. Our approach fills the research gap in low-resolution video HPE.

\section{Method}

\subsection{Overall Architecture}
\begin{figure}[htb]
\centering
\includegraphics[width=0.95\textwidth]{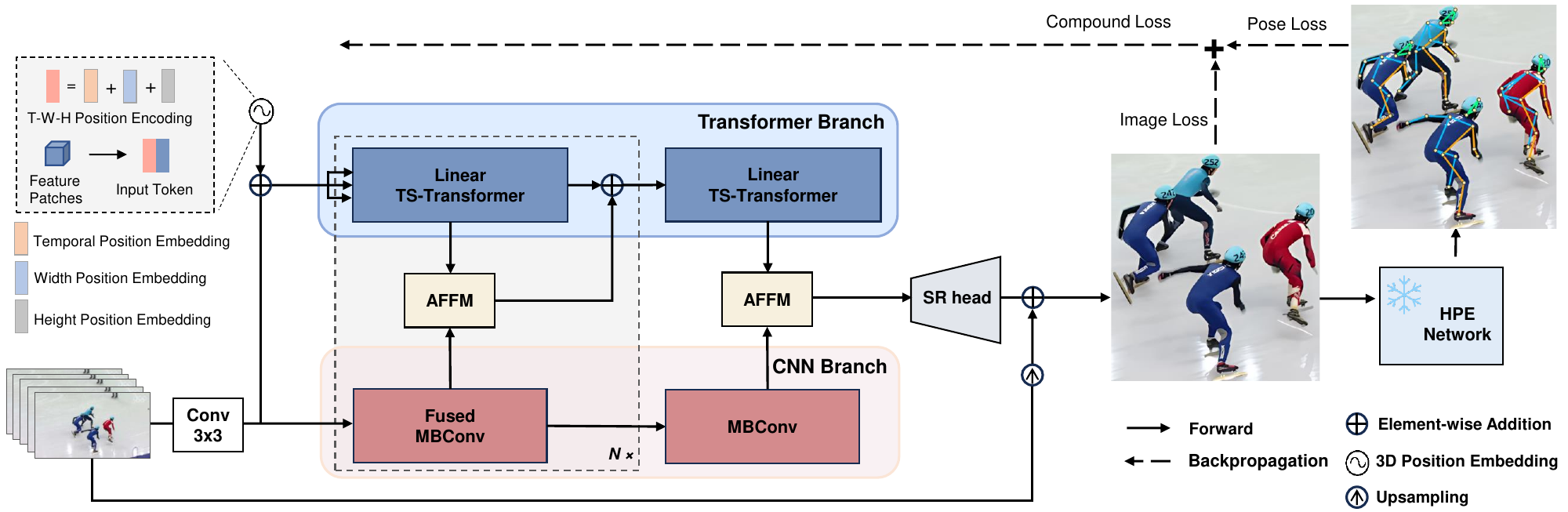}
\caption{Overview of the STAR-Pose framework. It consists of three core components: (a) spatial-temporal transformer branch with linear attention; (b) parallel CNN branch; (c) adaptive feature fusion module (AFFM). } \label{fig1}
\end{figure}

STAR-Pose employs a dual-branch architecture to collaboratively capture global spatial-temporal dependencies and local texture details from  low-resolution videos. As shown in Figure \ref{fig1}, Given $T$ consecutive low-resolution frames ($T=5$) as input, $\{I_t\}_{t=1}^T \in \mathbb{R}^{T \times 3 \times H^{'} \times W^{'}}$, initial features $F_0 \in \mathbb{R}^{T \times C \times H \times W}$ are extracted via using  $3\times3$ convolution and processed through three core components:

\begin{enumerate}
    \item \textbf{Linear Spatial-Temporal Transformer Branch}: Features are first partitioned into non-overlapping 3D patches through unfold operations and transformed to token sequences. After 3D positional encoding, LeakyReLU-based linear attention\cite{LinearTransformer1} models long-range spatial-temporal dependencies with $\mathcal{O}(N)$ complexity.
    \item \textbf{Parallel CNN Branch}: Lightweight convolutions extract local details including edges and textures, compensating for the expressive limitations of linear attention.
    \item \textbf{Adaptive Fusion Module}: A channel attention mechanism dynamically fuses features from both branches at multiple processing stages, enabling complementary integration of local and global information.
\end{enumerate}

Fused features are upsampled via pixel shuffle\cite{SR2} and fed into replaceable HPE networks (e.g., HRNet\cite{HRNet}) for keypoint prediction. During end-to-end training, a pose-aware compound loss guides the SR process to preferentially reconstruct details and sharp body contours that are most beneficial for keypoint localization.

\subsection{Efficient ST-Transformer with LeakyRelu Linear Attention}
Unlike traditional ViT\cite{VIT} that divides 2D patches on single frames, we perform division in feature space and encode spatial-temporal positional information. As shown in Fig.~\ref{fig1}, for feature patch at timestep $t$ and spatial position $(h, w)$, 3D positional encoding is:

\begin{equation}
\text{PE}_{\text{3D}}(t, w, h) = \text{Concat}\left(\left[\text{PE}_t(t),\, \text{PE}_w(w),\, \text{PE}_h(h)\right]\right)\cdot W_{pos}
\label{equ1}
\end{equation}
where $\text{PE}_t$, $\text{PE}_h$, and $\text{PE}_w$ represent sinusoidal position encodings for temporal and spatial dimensions, and $W_{pos}$ is a learnable projection matrix. This T-W-H positional embedding enables precise capture of spatial-temporal relationships among each feature map.

Standard self-attention in Transformers faces quadratic computational complexity of $\mathcal{O}(N^2)$  for long sequence data such as video. Linear attention\cite{LinearTransformer1} reduces the complexity to $\mathcal{O}(N)$ by redefining the Q-K-V interactions. Inspired by this, We propose an improved linear attention mechanism based on LeakyReLU for efficient long-range dependency modeling.

Given input sequence $X \in \mathbb{R}^{N \times F}$ of $N$ feature vectors of dimension $F$, a learnable linear mapping with $W_Q, W_K, W_V \in \mathbb{R}^{F \times D}$ is used to obtain matrices $Q, K, V \in \mathbb{R}^{N \times D}$, where $D$ represents the feature dimension. Self-attention computation can be expressed as:
\begin{equation}
O_{i}=\sum_{j=1}^{N} \frac{\operatorname{Sim}\left(Q_{i}, K_{j}\right)}{\sum_{j=1}^{N} \operatorname{Sim}\left(Q_{i}, K_{j}\right)} V_{j}
\label{equ2}
\end{equation}
where $Sim(\cdot,\cdot)$ is similarity function. When ${Sim}(Q, K)=\exp \left(Q K^{T} / \sqrt{d}\right)$, it represents the original Softmax-based attention. The computational cost of $\mathcal{O}(N^2)$ from this similarity function, mainly comes from the matrix multiplication between $Q$ and $K$\cite{TransSurvey1}. 

Linear attention introduces a linear mapping function  $\phi (·)$, which redefines the similarity calculation as $\text{Sim}(Q_i, K_j)=\phi(Q_i) \cdot \phi(K_j)^T$. Existing approaches\cite{LinearTransformer1,LinearTransformer3} typically employ $\phi=ELU(x)+1$ or $\phi=ReLU(x)$,  but suffer from computationally expensive exponential operations or ``Dying ReLU'' issue\cite{DyingRelu}. To balance efficiency and expressive power, we propose Shifted and Clipped LeakyReLU (SCLeakyReLU) as mapping function $\phi(\cdot)$:

\begin{equation}
\phi(x)=\text { SCLeakyReLU }(x)=\left\{\begin{array}{ll}
x+\delta  & \text { if } x>0 \\
\alpha x +\delta & \text { if } -1/\alpha \leq x \leq 0 \\
0 & \text { if } x< -1/\alpha
\end{array}\right.
\end{equation}
The truncation rate $\alpha$ is learnable (initialized as 0.01) and the offset $\delta = 1$. SCLeakyReLU retains partial negative directional information while avoiding ``Dying ReLU'' issue, and eliminating computationally expensive exponential computations.

\begin{figure}[htb]
\centering
\includegraphics[width=0.95\textwidth]{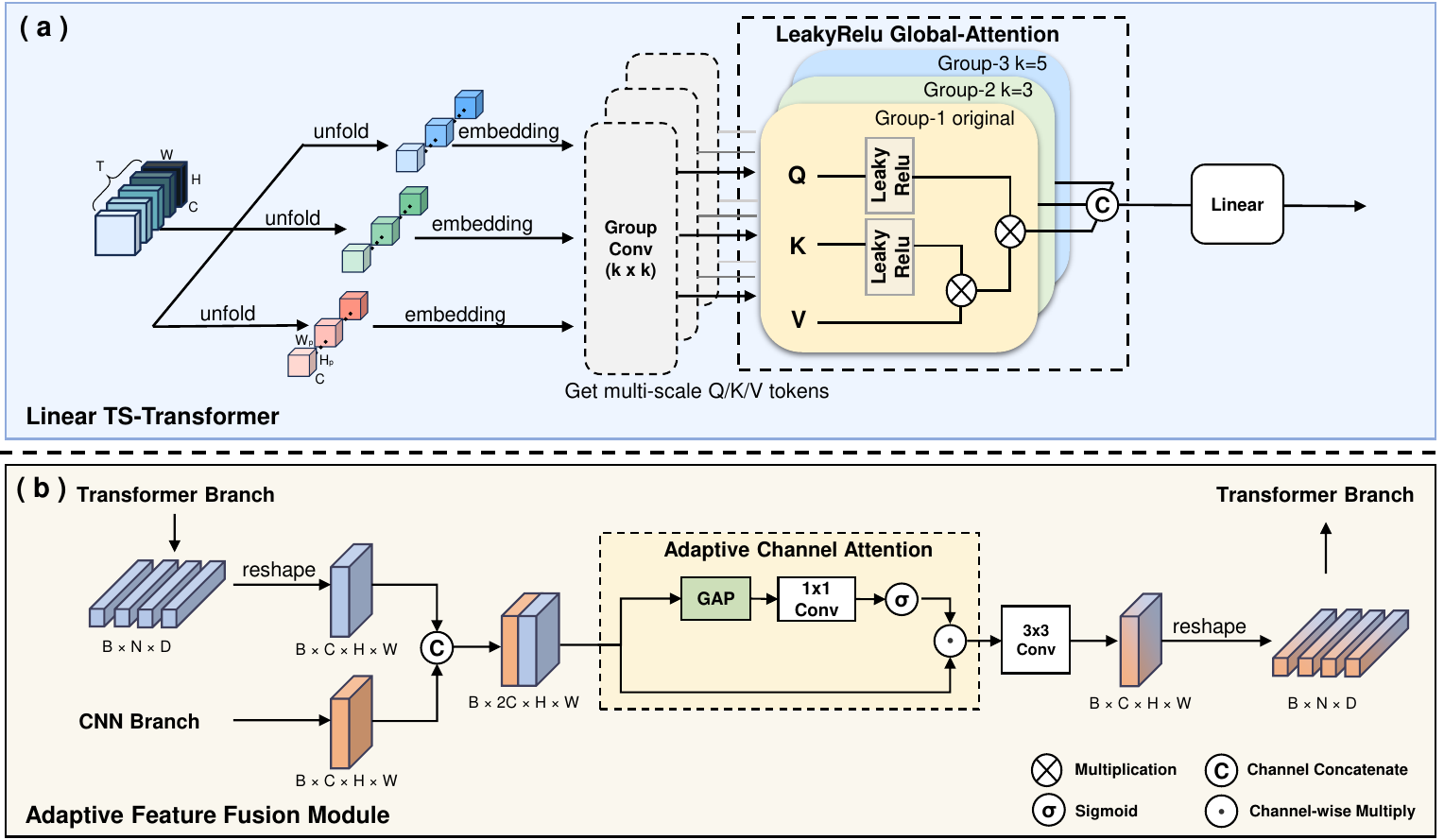}
\caption{Details of key components in STAR-Pose. (a) LeakyReLU-based linear spatial-temporal transformer architecture; (b) adaptive feature fusion module (AFFM).} \label{fig2}
\end{figure}

Linear attention further exploits the associativity of matrix multiplication to transform Equation(\ref{equ2}), leading to an efficient computational structure, as shown ``LeakyReLU Global Attention'' in Fig.~\ref{fig2}(a). This structure requires first calculating two global statistics: the accumulation term of the key-value interaction $ (\sum_{j=1}^{N} \operatorname{SCLeakyReLU}\left(K_{j}\right)^{T} V_{j} )\in \mathbb{R}^{D \times D}$ (in our case) and the normalization factor $ (\sum_{j=1}^{N} \operatorname{SCLeakyReLU}\left(K_{j}\right)^{T})\in \mathbb{R}^{D \times 1}$. These pre-computed terms can be shared and reused across all query positions $Q_i$, thus reducing the complexity from $\mathcal{O}(N^2 \cdot D)$ to $\mathcal{O}(N \cdot D^2)$. When $N>>D$ (which holds for video sequences), this actually achieves linear complexity $\mathcal{O}(N)$ respect to the sequence length, making long sequence modeling efficient and feasible.

Although linear attention offers significant advantages in computational efficiency, it has certain limitations in model capacity\cite{TransSurvey1}. We adopted two strategies to mitigate this limitation: (1) Following the multi-scale design proposed in \cite{LinearTransformer3}, we construct three parallel token groups before self-attention computation, including original input tokens and those processed by 3×3 and 5×5 depthwise convolutions. Different convolutional head aggregate nearby information at multiple spatial scales. (2) We design a parallel lightweight CNN branch as a supplementary path focus on local feature extraction.

This design maintains computational efficiency while providing rich spatial-temporal representations for super-resolution reconstruction.

\subsection{Lightweight CNN Branch with Adaptive Feature Fusion}
While spatial-temporal Transformers excel at long-range dependencies, CNNs remain superior for capturing local spatial patterns and texture details. We design parallel lightweight CNN branch with adaptive fusion to leverage these complementary strengths and further mitigate the limitations of linear attention.

For computational efficiency, CNN branches employ Inverted Residual Blocks (MBConv) from MobileNetV3\cite{MobileNetV3} and Fused-MBConv from EfficientNetV2\cite{EfficientNetV2}. Fused-MBConv merges expansion and depthwise convolutions into single standard convolutions at shallow layers. Specifically, we use Fused-MBConv in the first three stages and use MBConv in subsequent stages, all with 3×3 kernels. This design enhances multi-scale local features learning with little parameter overhead.

Due to structural differences between sequence-like token embeddings produced by Transformer and spatially structured feature maps generated by CNN, direct fusion through simple addition is infeasible. To dynamically integrate complementary information from both branches, we propose an Adaptive Feature Fusion Module (AFFM) , as shown in Fig.~\ref{fig2}(b).

For the $i$-th fusion stage, given CNN output $F_{i}^{C}\in \mathbb{R}^{B\times C\times H\times W}$ and Transformer output $F_{i}^{T}\in \mathbb{R}^{B\times N \times D}$, we reshape $F_{i}^{T}$ into an image-like feature map $\tilde{F}_{i}^{T}\in \mathbb{R}^{B\times C\times H\times W}$ to match the spatial organization of CNN features. Then concatenate these aligned features along channel dimension:
\begin{equation}
F_{i}^{TC}=Concat(Reshape(F_{i}^{T}),F_{i}^{C})\in \mathbb {R}^{B \times 2C \times H \times W}
\end{equation}
Subsequently, channel attention dynamically adjusts the importance of each channel, followed by 3×3 convolution for dimension reduction, resulting in an initially fused feature $\tilde{F}_{i}^{M}$:
\begin{equation}
\tilde{F}_{i}^{M} = Conv_{3\times3}[F_{i}^{TC} \odot \sigma(W_1 \cdot GAP(F_{i}^{TC}))] \in \mathbb{R}^{B \times C \times H \times W}
\end{equation}
where $\sigma$ is Sigmoid function, $W_1$ is a linear transformation, and $GAP$ represents global average pooling. After reshaping $\tilde{F}_{i}^{M}$ back into token format ${F}_{i}^{M} \in \mathbb{R}^{B \times N \times D}$, a learnable parameter $\beta$ (initialized as 0) adaptively integrates the complementary information into the input of next Transformer block:
\begin{equation}
{F}_{i+1}^{T,in} = {F}_{i}^{T,out} + \beta \times {F}_{i}^{M}
\end{equation}
The CNN branch remains independent to focus on purely local texture extraction, while multi-stage fusion complements global modeling advantage of Transformer with local spatial detail advantage of CNN.
\subsection{Pose-Guided Compound Loss}
SR models that optimizing only quality of pixel reconstruction tend to focus on improving overall image similarity, which may not achieve optimal pose estimation performance. Therefore, we introduce a pose-guided compound loss that, while pursuing pixel-level image fidelity, leverages pose information to guide the SR processing beneficial for human pose estimation. The compound loss $\mathcal{L}$ is defined as:
\begin{equation}
\mathcal{L}=\underbrace{\left\|I_{S R}-I_{H R}\right\|_{1}}_{\mathcal{L}_{\text {pixel }}}+\lambda \underbrace{\|\hat{H}-H\|_{2}^{2}}_{\mathcal{L}_{\text {pose }}}
\label{equ10}
\end{equation}
where $\mathcal{L}_{\text{pixel}}$ is L1 loss between SR image $I_{SR}$ and ground-truth high-resolution image $I_{HR}$, which is used to restore high-frequency details in image. $\mathcal{L}_{\text{pose}}$ is MSE loss between predicted pose heatmap $\hat{H}$(from pre-trained HPE model) and ground-truth heatmap $H$. Weight factor $\lambda$ is used to balance the contributions of two loss terms.

$\mathcal{L}_{\text{pose}}$ acts like a task-specific perceptual loss using human pose as a high-level structural prior. From an optimization theory perspective, this creates complementary gradient flows in model's parameter space: when gradients derived from human structure and pixel-level are aligned, they reinforce each other to accelerate model convergence; when their directions conflict, pose-related gradient prevents model from over-optimizing irrelevant details. Such gradient interaction is particularly beneficial for handling low-quality inputs, as structural cues are often easier to extract from low-resolution video than subtle texture detail.
\begin{figure}
\centering
\includegraphics[width=0.95\textwidth]{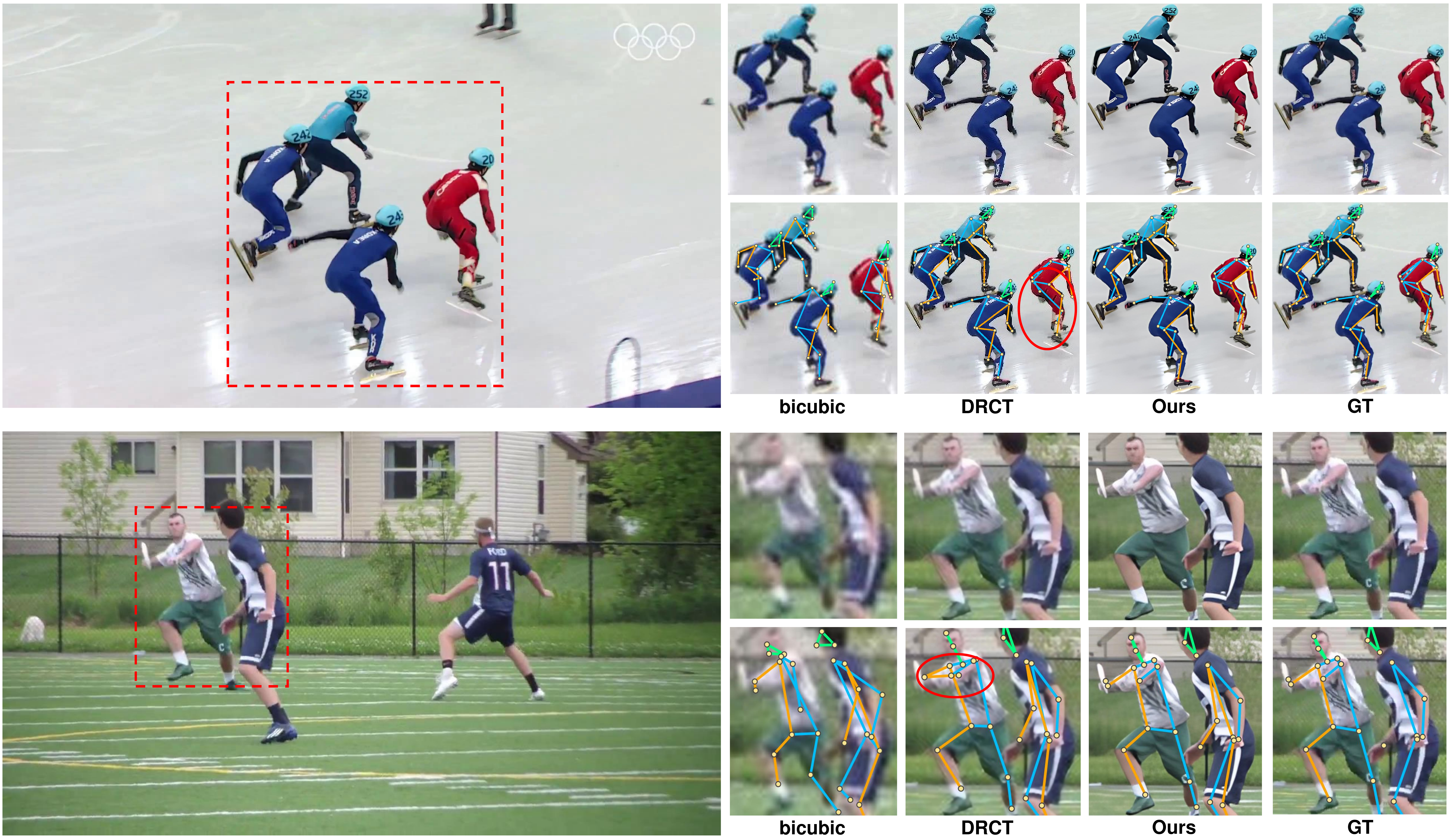}
\caption{ Qualitative comparison of super-resolution outputs and corresponding pose estimation results across different methods. } \label{fig3}
\end{figure}
\section{Experiments}
\subsection{Experimental Settings}
\subsubsection{DataSet and Evaluation Metrics}
We evaluate our method on two large-scale video human pose estimation datasets: PoseTrack2018\cite{PoseTrack18} and PoseTrack2021\cite{PoseTrack21}. These datasets encompass diverse real-world scenarios including challenging dynamic movements and crowded multi-person interactions, and contain 153,615 and 177,164 pose annotations for 15 human keypoints, respectively. For evaluation metrics, we utilize mean Average Precision (mAP) and mean Average Recall (AR) to assess keypoint localization performance. Additionally, we report image reconstruction quality with Peak Signal-to-Noise Ratio (PSNR) and Structural Similarity Index Measure (SSIM), along with each model's inference time per frame.

\subsubsection{Implementation Details}
To systematically simulate low-resolution scenarios, we apply Gaussian blurring to original videos, followed by 2×/4× bicubic downsampling, generating two distinct resolution levels: low and ultra-low. Single-person bounding boxes are adjusted to fixed resolutions of 128x96 and 64×48 based on provided scale and center annotations of PoseTrack. Data augmentation includes random rotation (±30°), scaling (0.75-1.25), horizontal flipping and photometric distortion. Regarding network configuration, the spatial-temporal Transformer comprises 8 layers, each with 8 attention heads and hidden dimension of 256. The parallel CNN branch incorporates 3 Fused-MBConv blocks and 5 MBConv blocks, all with kernel size of 3×3. We adopt the pre-trained HRNet-W48\cite{HRNet} as pose estimation backbone, with frozen parameters during training to ensure fair comparison. The model is trained for 300 epochs with 
a batch size 32 using the Adam optimizer, setting initial learning rate to 1e-3, subsequently decaying to 1e-4 and 1e-5 at epoch 170 and 260, respectively. All experiments are conducted using PyTorch on an NVIDIA RTX 3090 GPU.

\begin{table}[htbp]
\caption{Quantitative comparison of HPE performance on PoseTrack2018 dataset.}
\label{tab1}
\centering
\small
\setlength{\tabcolsep}{4pt} 
\resizebox{\linewidth}{!}{
\begin{tabular}{@{} l l  S[table-format=3.1]  S[table-format=2.1]  | *{4}{S[table-format=2.1]}  c @{}}
\toprule
\textbf{Method} & \textbf{Backbone} & \textbf{\#Params (M)} & \textbf{GFLOPs} & \textbf{AP@0.5} & \textbf{AP@0.75} & \textbf{mAP} & \textbf{AR} & \textbf{Time (ms)} \\
\midrule
\multicolumn{9}{c}{\textbf{128$\times$96}} \\
\midrule
HRNet\cite{HRNet} (baseline)  & HRNet-W48 & 63.6   & 3.65 & 69.3 & 52.8 & 57.6 & 61.2 &  \underline{45.2} \\
Bicubic+HRNet\cite{HRNet}     & HRNet-W48 & 63.6   & 3.65 & 70.5 & 54.0 & 58.9 & 62.5 &  46.1 \\
UDP\cite{UDP}                 & HRNet-W48 & 63.8   & 3.77 & 72.4 & 55.8 & 60.5 & 64.0 &  46.8 \\
DCPose\cite{DCPose}           & HRNet-W48 & 65.2   & 4.75 & 73.6 & 56.7 & 61.8 & 65.3 &  53.2 \\
SRPose\cite{SRHeatMap}        & HRNet-W32 & 29.9   & 3.14 & 75.8 & 61.8 & 64.3 & 67.9 &  \textbf{37.5}\\
DSTA\cite{DSTA}               & ViT-H     & 63.9   & 3.95 & \underline{77.9} & \underline{63.8} & \underline{68.2} & \underline{71.3} &  48.7\\
SDBVSR\cite{SR19}+HRNet       & HRNet-W48 & 82.2   & 12.19 & 75.8 & 63.1 & 65.2 & 68.7 & 311.8 \\
DRCT\cite{SR21}+HRNet         & HRNet-W48 & 74.1   & 11.57 & 77.4 & 64.9 & 67.1 & 69.3 & 201.7 \\
\textbf{STAR-Pose}            & HRNet-W48 & 72.7   & 6.75  & \textbf{79.7} & \textbf{67.3} &\textbf{71.6} & \textbf{73.8} &  85.3\\
\midrule
\multicolumn{9}{c}{\textbf{64$\times$48}} \\
\midrule
HRNet\cite{HRNet} (baseline)  & HRNet-W48 & 63.6   & 0.91 & 56.4 & 39.2 & 44.3 & 48.5 &  \underline{18.1} \\
Bicubic+HRNet\cite{HRNet}     & HRNet-W48 & 63.6   & 0.91 & 58.6 & 41.5 & 46.8 & 50.7 &  18.5\\
UDP\cite{UDP}                 & HRNet-W48 & 63.8   & 0.97 & 60.1 & 43.1 & 48.3 & 52.4 &  19.2 \\
DCPose\cite{DCPose}           & HRNet-W48 & 65.2   & 1.21 & 61.0 & 44.2 & 49.6 & 53.9 &  22.8 \\
SRPose\cite{SRHeatMap}        & HRNet-W32 & 29.9   & 0.79 & 64.2 & 49.8 & 53.9 & 57.5 &  \textbf{15.6}\\
DSTA\cite{DSTA}               & ViT-H     & 63.9   & 0.99 & 66.5 & 52.1 & 56.2 & 59.9 &  20.6 \\
SDBVSR\cite{SR19}+HRNet       & HRNet-W48 & 82.2   & 3.35 & 70.1 & 57.8 & 60.5 & 63.9 & 142.5 \\
DRCT\cite{SR21}+HRNet         & HRNet-W48 & 74.1   & 3.02 & \underline{71.8} & \underline{59.5} & \underline{62.2} & \underline{65.5} & 93.1  \\
\textbf{STAR-Pose}            & HRNet-W48 & 72.7   & 1.71 & \textbf{75.2} & \textbf{63.8} & \textbf{67.4} & \textbf{69.9} &  32.4 \\
\bottomrule
\end{tabular}}
\end{table}

\subsection{Quantitative Comparison and Qualitative Analysis}
Table~\ref{tab1} presents a performance comparison on the PoseTrack 2018 validation set. Under the low-resolution (128×96) condition, STAR-Pose achieves 71.6\% mAP, surpassing the second-best method DSTA\cite{DSTA} by 3.4\%. Under the extremely low resolution (64×48), STAR-Pose demonstrates remarkable robustness, maintaining 67.4\% mAP, leading all baseline methods by at least 5.2\%. 

Note that the VSR+HPE cascaded approaches outperform pure HPE methods under extremely low resolution, demonstrating greater resilience against performance degradation. However, this comes at the cost of substantially increased computational overhead. For instance, SDBVSR\cite{SR19}+HRNet requires 142.5ms and DRCT\cite{SR21}+HRNet needs 93.1ms per frame. In contrast, STAR-Pose achieves only 32.4ms per frame, providing significant efficiency improvements while maintaining superior accuracy. Notably, although requiring higher computational resources and inference time than the lightweight backbone HRNet-W32 of SRPose\cite{SRHeatMap}, the substantial performance gains under extremely low-resolution conditions validate our task-driven SR approach.

\begin{table}[htbp]
\caption{Quantitative comparison of HPE performance on PoseTrack2021 dataset.}
\label{tab2}
\centering
\small
\setlength{\tabcolsep}{4pt} 
\resizebox{\linewidth}{!}{
\begin{tabular}{@{} l l S[table-format=3.1] S[table-format=2.1]  | *{4}{S[table-format=2.1]} c @{}}
\toprule
\textbf{Method} & \textbf{Backbone} & \textbf{\#Params (M)} & \textbf{GFLOPs} & \textbf{AP@0.5} & \textbf{AP@0.75} & \textbf{mAP} & \textbf{AR} & \textbf{Time (ms)} \\
\midrule
\multicolumn{9}{c}{\textbf{128$\times$96}} \\
\midrule
HRNet\cite{HRNet} (baseline)  & HRNet-W48 & 63.6   & 3.65  & 67.5 & 50.9 & 55.8 & 59.4 &  \underline{47.3}\\
Bicubic+HRNet\cite{HRNet}     & HRNet-W48 & 63.6   & 3.65  & 68.7 & 52.2 & 57.1 & 60.7 &  48.2\\
UDP\cite{UDP}                 & HRNet-W48 & 63.8   & 3.77 & 70.6 & 54.0 & 58.7 & 62.2 &  49.0 \\
DCPose\cite{DCPose}           & HRNet-W48 & 65.2   & 4.75  & 71.8 & 55.0 & 60.1 & 63.5 &  55.8\\
SRPose\cite{SRHeatMap}        & HRNet-W32 & 29.9   & 3.14  & 74.1 & 60.1 & 62.8 & 66.2 &  \textbf{39.1}\\
DSTA\cite{DSTA}               & ViT-H     & 63.9   & 3.95  & \underline{76.1} & \underline{62.8} & \underline{65.7} & \underline{68.9} &  50.5\\
SDBVSR\cite{SR19}+HRNet       & HRNet-W48 & 82.2   & 12.19  & 74.2 & 60.3 & 63.1 & 66.4 & 323.5\\
DRCT\cite{SR21}+HRNet         & HRNet-W48 & 78.1   & 11.57 & 75.8 & 61.9 & 64.7 & 68.0 & 210.4 \\
\textbf{STAR-Pose}            & HRNet-W48 & 72.7   & 6.75  & \textbf{78.9} & \textbf{65.5} & \textbf{69.1} & \textbf{71.4} & 88.1 \\
\midrule
\multicolumn{9}{c}{\textbf{64$\times$48}} \\
\midrule
HRNet\cite{HRNet} (baseline)  & HRNet-W48 & 63.6   & 0.91 & 54.8 & 37.2 & 42.1 & 46.3 &  \underline{18.9} \\
Bicubic+HRNet\cite{HRNet}     & HRNet-W48 & 63.6   & 0.91  & 57.0 & 39.6 & 44.7 & 48.6 &  19.3\\
UDP\cite{UDP}                 & HRNet-W48 & 63.8   & 0.97 & 58.5 & 41.2 & 46.2 & 50.2 &  20.1 \\
DCPose\cite{DCPose}           & HRNet-W48 & 65.2   & 1.21 & 59.4 & 42.3 & 47.5 & 51.7 &  23.2 \\
SRPose\cite{SRHeatMap}        & HRNet-W32 & 29.9   & 0.79 & 62.6 & 47.8 & 52.4 & 56.3 &  \textbf{16.3} \\
DSTA\cite{DSTA}               & ViT-H     & 63.9   & 0.99  & 64.8 & 50.5 & 54.8 & 58.6 &  19.8 \\
SDBVSR\cite{SR19}+HRNet       & HRNet-W48 & 82.2   & 3.35  & 68.3 & 55.1 & 58.3 & 61.9 & 147.3\\
DRCT\cite{SR21}+HRNet         & HRNet-W48 & 78.1   & 3.02 & \underline{70.0} & \underline{56.8} & \underline{59.9} & \underline{63.5} & 95.6 \\
\textbf{STAR-Pose}            & HRNet-W48 & 72.7   & 1.71  & \textbf{74.5} & \textbf{62.7} & \textbf{65.5} & \textbf{67.9} & 33.9\\
\bottomrule
\end{tabular}}
\end{table}

Table~\ref{tab2} shows results on challenging PoseTrack 2021 validation set. Despite an overall performance decline compared to PoseTrack 2018, STAR-Pose maintains significant advantages:  69.1\% mAP at 128×96 (3.4\% higher) and 65.5\% mAP at 64×48 (5.6\% higher), while maintaining 3–4× faster inference speed than cascaded approaches.

To further analyze image reconstruction quality, we evaluate various super-resolution models on 4× downsampled PoseTrack21 dataset, as shown in Table~\ref{tab3}. Although STAR-Pose achieves slightly lower pixel-level similarity than general SR models, it attains a competitive PSNR of 30.86 dB at significantly lower inference time (14.3 ms). This demonstrates a balanced trade-off between reconstruction quality and computational efficiency, validating STAR-Pose’s task-driven design prioritizing restoration of semantically relevant structures over pixel-perfect fidelity.

Qualitative visual comparisons in Fig.~\ref{fig3} show that STAR-Pose reconstructs clearer human contours and sharper wrinkle textures compared to bicubic interpolation or generic SR methods lacking human structure awareness. Additionally, our method effectively separates foreground from background. These improvements enable more accurate localization of critical keypoints such as wrists and elbows, and markedly reduce misidentifications between left and right limbs, validating the effectiveness of task-driven design.

\subsection{Ablation Study}
\subsubsection{Impact of Key Components}
To validate component contributions, we conducted a systematic ablation study on PoseTrack2018 validation set. Table~\ref{tab4} shows the impact of progressively removing key modules, ``w/o'' denotes the removal of corresponding component:

(a) Removing temporal context modeling caused the most significant performance drop, with mAP decreasing by 4.5\% and 6.7\% at 128×96 and 64×48 resolutions respectively. This decline is more severe under ultra-low resolution, highlighting temporal context's critical role in recovering human structural information from heavily degraded video input. 

(b) Replacing  SCLeakyReLU with standard ReLU in linear attention mapping function resulted in mAP drops of 1.8\% and 1.1\%, plus 1.21 dB PSNR decrease. These results indicate the importance of non-zero negative region in LeakyReLU for maintaining feature flow within linear attention, particularly for capturing fine details from low-resolution inputs. 

\begin{table}[htbp]
\caption{Comparison of SR performance on 4× downsampled PoseTrack2018 dataset.}
\label{tab3}
\centering
\footnotesize
\scalebox{0.9}{
\begin{tabular}{@{} l S[table-format=2.1] |S[table-format=2.2] S[table-format=1.3]  S[table-format=3.2] @{}}
\toprule
\textbf{Method} & \textbf{Params (M)} & \textbf{PSNR (dB)} & \textbf{SSIM} & \textbf{Time (ms)} \\
\midrule
Bicubic                 & {--} & 28.21 & 0.835  &   {--} \\
SwinIR\cite{SR8}        & 11.9  & 29.45 & 0.861 & 78.5 \\
BasicVSR++\cite{SR12}   &  7.3 & 30.58 & 0.883  &  45.2 \\
VRT\cite{SR13}          & 35.6  & 31.24 & 0.897 & 165.7 \\
RVRT\cite{SR14}         & 10.8 & 31.63 & 0.903  & 118.3 \\
HAT\cite{SR16}          & 20.8  & 31.95 & 0.907 & 125.6 \\
SDBVSR\cite{SR19}       & 20.8  & 32.17 & 0.909 & 124.4 \\
DRCT\cite{SR21}         & 14.1  & \textbf{32.52} & \textbf{0.914} & 74.9 \\
\textbf{STAR-Pose (Ours)} &  7.9 & 30.86 & 0.887 & \textbf{14.3} \\
\bottomrule
\end{tabular}}
\end{table}

(c) Removal of the CNN branch and the adaptive fusion module led to a drop of 3.3 to 4.5\% mAP and a decrease of 2.3 dB PSNR. This demonstrates that local texture features extracted by CNN effectively complement Transformer's global modeling capability, with their synergy essential for performance improvement.

(d) Removing pose-guided compound loss and training with only pixel loss slightly improved PSNR by 0.26 dB. However, pose estimation performance declined by 1.5–1.7\% mAP. This supports our claim that optimizing solely for pixel-level reconstruction doesn't guarantee optimal downstream-task performance. Task-driven supervision effectively guides the network toward feature representations more valuable for pose estimation.

Comprehensive analysis reveals that STAR-Pose components form a tightly integrated framework achieving efficiency-accuracy balance: spatial-temporal modeling provides motion continuity cues, linear attention ensures computational efficiency, CNN branch supplements local detail, and task-driven loss guarantees HR images optimized for pose estimation task.

\begin{table}[htbp]
\caption{Ablation analysis showing the contribution of each component in STAR-Pose.}
\label{tab4}
\centering
\footnotesize
\scalebox{0.9}{
\begin{tabular}{@{} >{\bfseries}l 
                  S[table-format=2.1] 
                  S[table-format=2.1] 
                  S[table-format=2.2] 
                  S[table-format=1.3] 
                  S[table-format=2.1] @{}}
\toprule
\textbf{Method} & 
\multicolumn{1}{c}{\textbf{mAP\tiny{(128×96)}}} & 
\multicolumn{1}{c}{\textbf{mAP\tiny{(64×48)}}} & 
\multicolumn{1}{c}{\textbf{PSNR}} & 
\multicolumn{1}{c}{\textbf{SSIM}} & 
\multicolumn{1}{c}{\textbf{Time}} \\
\midrule
STAR-Pose (Full) & 71.6 & 67.4 & 30.86 & 0.887 & 32.4 \\
\addlinespace[0.2cm]
(a) w/o Temporal Context & 67.1 & 60.7 & 28.42 & 0.841 & 24.6 \\
(b) w/o LeakyReLU       & 69.8 & 66.3 & 29.65 & 0.864 & 32.2 \\
(c) w/o CNN Branch      & 68.3 & 62.9 & 28.56 & 0.845 & 29.5 \\
(d) w/o Compound Loss   & 70.1 & 65.7 & 31.12 & 0.892 & 32.4 \\
\bottomrule
\end{tabular}}
\end{table}

\subsubsection{Sensitivity to Loss Balancing Parameter}
The balancing parameter $\lambda$ in compound loss controls relative importance between pixel reconstruction and pose optimization. Table~\ref{tab5} presents performance variations under different $\lambda$ settings, revealing a clear trade-off trend: when $\lambda$ is small (0.01–0.1), the model focuses more on pixel-level reconstruction, yielding high PSNR/SSIM but limited keypoint localization accuracy. As $\lambda$ increases, the model gradually shifts toward pose estimation optimization, improving mAP correspondingly. At $\lambda = 10$, the model achieves optimal balance, reaching peak mAP values (71.6\% and 67.4\% at both resolutions) while maintaining acceptable reconstruction quality (30.86 dB PSNR). However, when $\lambda$ is too high (e.g., 100), the model overemphasizes pose-related features at expense of overall image quality, which in turn degrades keypoint localization accuracy. 

This non-monotonic performance curve reflects the complexity of multi-objective optimization: effective task-driven SR must enhance task-relevant features while preserving reasonable visual fidelity. Based on experimental findings, we adopt $\lambda = 10$ as default configuration.

\begin{table}[htbp]
\caption{Sensitivity analysis of the compound loss balancing parameter $\lambda$}
\label{tab5}
\centering
\setlength{\tabcolsep}{4pt} 
\footnotesize
\scalebox{0.9}{
\begin{tabular}{@{} l *{8}{c} @{}}
\toprule
\multirow{2}{*}{\textbf{Parameter}} & \multicolumn{8}{c}{\textbf{Value of $\lambda$}} \\
\cmidrule(lr){2-9}
& {0.01} & {0.1} & {1} & {5} & {10} & {25} & {50} & {100} \\
\midrule
\textbf{mAP (128$\times$96)} & 70.3 & 70.6 & 70.9 & 71.3 & \textbf{71.6} & 71.2 & 70.5 & 69.8 \\
\textbf{mAP (64$\times$48)}  & 65.8 & 66.1 & 66.5 & 66.9 & \textbf{67.4} & 66.8 & 66.0 & 65.4 \\
\textbf{PSNR}         & \textbf{31.09} & 30.95 & 30.92 & 30.89 & 30.86 & 30.53 & 30.21 & 29.87 \\
\textbf{SSIM}         & \textbf{0.892} & 0.890 & 0.889 & 0.888 & 0.887 & 0.882 & 0.874 & 0.868 \\
\bottomrule
\end{tabular}}
\end{table}

\section{Conclusion}
In this work, we propose STAR-Pose, a novel spatial-temporal adaptive super-resolution framework designed for the challenging problem of human pose estimation in low-resolution videos. Our approach integrates a spatial-temporal Transformer with LeakyReLU-enhanced linear attention and a lightweight CNN branch, effectively capturing long-range temporal dependencies while maintaining computational efficiency. The pose-guided compound loss enables task-aware reconstruction, prioritizing structural features crucial for accurate keypoint localization over mere pixel-level fidelity.

Comprehensive experiments across multiple video HPE datasets demonstrate that STAR-Pose significantly outperforms state-of-the-art methods under extremely low-resolution conditions, achieving substantial improvements in both accuracy and inference speed. The systematic ablation studies validate the synergistic contribution of each component, confirming our design of balancing global temporal modeling with local detail enhancement.

Looking forward, we envision several promising research directions: exploring self-supervised learning paradigms to reduce dependency on paired high-low-resolution training data, and developing ultra-lightweight variants for edge deployment. These extensions will further broaden the applicability of task-driven super-resolution in real-world scenarios with diverse constraints.


\bibliographystyle{splncs04}
\bibliography{LaTeX2e_Proceedings_Templates/main}
%
%





\end{document}